# Efficient lattice field theory simulation using adaptive normalizing flow on a resistive memory-based neural differential equation solver

Meng Xu[1,2,3†], Jichang Yang[1,3†], Ning Lin[1,3], Qundao Xu[4], Siqi Tang[4], Han Wang[1], Xiaojuan Qi[1]*, Zhongrui Wang[2]*, Ming Xu[4]

[1]*Department of Electrical and Electronic Engineering, the University of Hong Kong, Hong Kong, China*

[2]*The School of Microelectronics, Southern University of Science and Technology, Shenzhen, China*

[3]*ACCESS - AI Chip Center for Emerging Smart Systems, InnoHK Centers, Hong Kong Science Park, Hong Kong, China*

[4]*School of Integrated Circuits, Huazhong University of Science and Technology, Wuhan, China*

[†]These authors contributed equally.

*Email: xjqi@hku.hk; wangzr@sustech.edu.cn;

**Abstract**

Lattice field theory (LFT) simulations underpin advances in classical statistical mechanics and quantum field theory, providing a unified computational framework across particle, nuclear, and condensed matter physics. However, the application of these methods to high-dimensional systems remains severely constrained by several challenges, including the prohibitive computational cost and limited parallelizability of conventional sampling algorithms such as hybrid Monte Carlo (HMC), the substantial training expense associated with traditional normalizing flow models, and the inherent energy inefficiency of digital hardware architectures. Here, we introduce a software-hardware co-design that integrates an adaptive normalizing flow (ANF) model with a resistive memory-based neural differential equation solver, enabling efficient generation of LFT configurations. Software-wise, ANF enables efficient parallel generation of statistically independent configurations, thereby reducing computational costs, while low-rank adaptation (LoRA) allows cost-effective fine-tuning across diverse simulation parameters. Hardware-wise, in-memory computing with resistive memory substantially enhances both parallelism and energy efficiency. We validate our approach on the scalar $\phi^4$ theory and the effective field theory of graphene wires, using a hybrid analog-digital neural differential equation solver equipped with a 180 nm resistive memory in-memory computing macro. Our co-design enables low-cost computation, achieving approximately 8.2-fold and 13.9-fold reductions in integrated autocorrelation time over HMC, while requiring fine-tuning of less than 8% of the weights via LoRA. Compared to state-of-the-art GPUs, our co-design achieves up to approximately 16.1- and 17.0-fold speedups for the two tasks, as well as 73.7- and 138.0-fold improvements in energy efficiency. Overall, our co-design paves the way for more efficient and affordable large-scale LFT simulations in high-dimensional physical systems.

# Introduction

Lattice field theory (LFT) provides a nonperturbative and first-principles framework for the investigation of classical statistical mechanics and quantum field theories[1-3]. It is a central tool across particle, nuclear, and condensed-matter physics, enabling quantitative exploration of phenomena such as topological defect formation, phase transitions, and nonequilibrium processes[4-6]. In classical statistical field theory, lattice methods allow for the numerical investigation of models such as the Ising model by sampling field configurations on a discrete lattice and studying their equilibrium properties[7-9]. In quantum field theory, LFT delivers precise determinations of hadron spectra and matrix elements in quantum chromodynamics, among other observables[10,11]. As a result, LFT continues to play a pivotal role in advancing our understanding of the fundamental laws governing nature.

Discretizing spacetime onto a lattice reformulates both classical and quantum field theories such that physical observables become high-dimensional integrals or sums over all field configurations[12]. The vast configuration space makes direct evaluation infeasible, so efficient stochastic sampling methods like Markov chain Monte Carlo (MCMC) are essential for sampling configurations according to the lattice action or Hamiltonian[13]. The hybrid Monte Carlo (HMC) algorithm is the main computational tool for LFT, but its high computational cost limits its use in large-scale or high-dimensional systems[14-17]. As lattice size and degrees of freedom increase, the computational burden grows rapidly, and large ensembles are needed to overcome statistical errors and autocorrelations, further limiting computational efficiency[18,19]. These challenges are compounded in high-dimensional or complex probability distributions, where HMC algorithms struggle to efficiently explore configuration space, manifesting as critical slowing down near phase transitions[20,21] and topological freezing in gauge theories[22]. In these regimes, autocorrelation times diverge and the algorithm can become trapped in fixed topological sectors, leading to highly correlated samples and biased estimates. As a result, achieving statistically independent measurements often requires prohibitively large ensembles, revealing fundamental limitations in the computational efficiency and scalability of HMC, and thus restricting the broader applicability of LFT. Moreover, the inherently sequential nature of HMC sampling poses challenges for fully leveraging the parallelism of modern computing architectures[23-25]. Although machine learning methods, such as flow-based generative models[26-30], have been proposed to accelerate LFT simulations, they introduce substantial training costs, as any change in the lattice action parameters requires retraining the model.

On the hardware side, both conventional HMC and machine learning-based approaches to LFT simulations fundamentally rely on traditional digital computing architectures, which intrinsically limit computational parallelism and energy efficiency[31]. A major issue is the von Neumann bottleneck[32-34], resulting from restricted data transfer speeds between off-chip memory and processing units. This bottleneck hampers overall computational efficiency due to the substantial time and energy overheads incurred by the frequent movement of large volumes of data. Additionally, while Moore's Law has historically driven progress in the semiconductor industry, the continual miniaturization of transistors is now approaching fundamental physical boundaries[35,36]. As a result, the rate of improvement predicted by Moore's Law has slowed and is anticipated to eventually halt, leading to a plateau in computational efficiency improvements achieved by conventional digital hardware systems.

To address these challenges, we present a software-hardware co-design that implements an adaptive normalizing flow (ANF) model on a resistive memory-based neural differential equation solver, enabling efficient generation of LFT configurations (**Fig. 1**). On the software front, the ANF model reduces the computational cost of LFT simulations by learning efficient transformations from a simple Gaussian distribution $r(x)$ to complex target distributions $p(\phi)$, where $p(\phi)$ is defined by the Euclidean action $S(\phi)$ (**Fig. 1a**). Unlike conventional MCMC approaches, in which each sample depends on the previous one, ANF generates statistically independent configurations as proposals[30]. Once trained with physics-guided action-based optimization (see **Supplementary Figure 1**), this independence drastically reduces autocorrelation times, mitigates critical slowing down[37,38], and facilitates ergodic sampling across topological sectors[39,40]. Furthermore, because each proposal is independent, ANF overcomes the inherent serial computation limitations of HMC, enabling efficient parallel sampling and substantially enhancing sampling efficiency, while maintaining accuracy comparable to HMC across a range of physical observables.

As depicted in **Fig. 1b**, ANF iteratively maps random samples $x$ drawn from $r(x)$ over multiple timesteps to field configurations $\phi$ that approximating $p(\phi)$ (see **Supplementary Figure 2**), with each timestep employing an invertible coupling layer $g_t^{-1}$ with shared weights (see **Methods**). As shown in **Fig. 1c**, $g_t^{-1}$ efficiently implements reversible transformations with tractable Jacobians[41,42] by partitioning inputs and transforming one subset conditioned on the other, which is realized via a mixer model based on low-rank adaptation (LoRA)[43,44] (see **Methods**). By integrating LoRA into the mixer blocks (**Fig. 1d**), the model enables flexible fine-tuning across lattice action parameters with minimal computational cost, while preserving strong generalization and transferability. In each inference step, the model projects each patch into the feature space using trainable patch embedding layers, alternately updates patch and channel representations through sequential mixer blocks, and finally computes the output via output embedding and regression layers (see **Methods**).

On the hardware front, we developed a hybrid analog-digital neural differential equation solver based on resistive memory, which further accelerates the ANF model. As illustrated in **Fig. 1e**, the analog part performs the matrix multiplications of primary neural networks, including the patch embedding layers, mixer blocks, output embedding layers, and regression layers, thereby encompassing the majority of the mixer model's weights (see **Methods**). By harnessing the efficient in-memory computing capabilities of resistive memory[45-57], computation and data storage are performed simultaneously on the analog part. This approach helps to overcomes the von Neumann bottleneck and markedly improves both parallelism and energy efficiency. The digital part handles a small subset of LoRA weights and other operations. Its high precision and flexible programmability enable compensation for resistive memory programming errors (see **Supplementary Figure 3**) and support fine-tuning across various LFT parameters without requiring analog part changes.

In this article, we demonstrate our approach on two representative LFT simulation tasks: scalar $\phi^4$ theory and effective field theory (EFT) of graphene wire[58-60], using a hybrid analog-digital neural differential equation solver equipped with a 180 nm resistive memory in-memory computing macro. On the software side, our ANF model reduces integrated autocorrelation time by up to approximately 8.2-fold in scalar $\phi^4$ theory and 13.9-fold in EFT of graphene wire compared to HMC, enabling the generation of statistically independent configurations with much lower computational cost. Notably, the digital LoRA weights account for only approximately 4.0% and 5.6% of the total ANF model weights in the two experiments, enabling low-cost mitigation of analog errors and facilitating efficient, adaptive adjustment to different lattice action parameters.

On the hardware side, leveraging the efficient in-memory computing of the analog resistive memory to perform approximately 99.2% and 99.6% of ANF's the multiply-accumulate (MAC) operations during inference in the two tasks, our hybrid analog-digital solver achieves up to 16.1-fold and 17.3-fold increases in average computational speed and 73.7-fold and 138.0-fold improvements in average energy efficiency, respectively, compared to an Nvidia RTX 5090 GPU. Overall, by leveraging resistive memory-based co-design, we address key computational challenges in large-scale physical systems and demonstrate substantial advantages over conventional LFT computational paradigms.

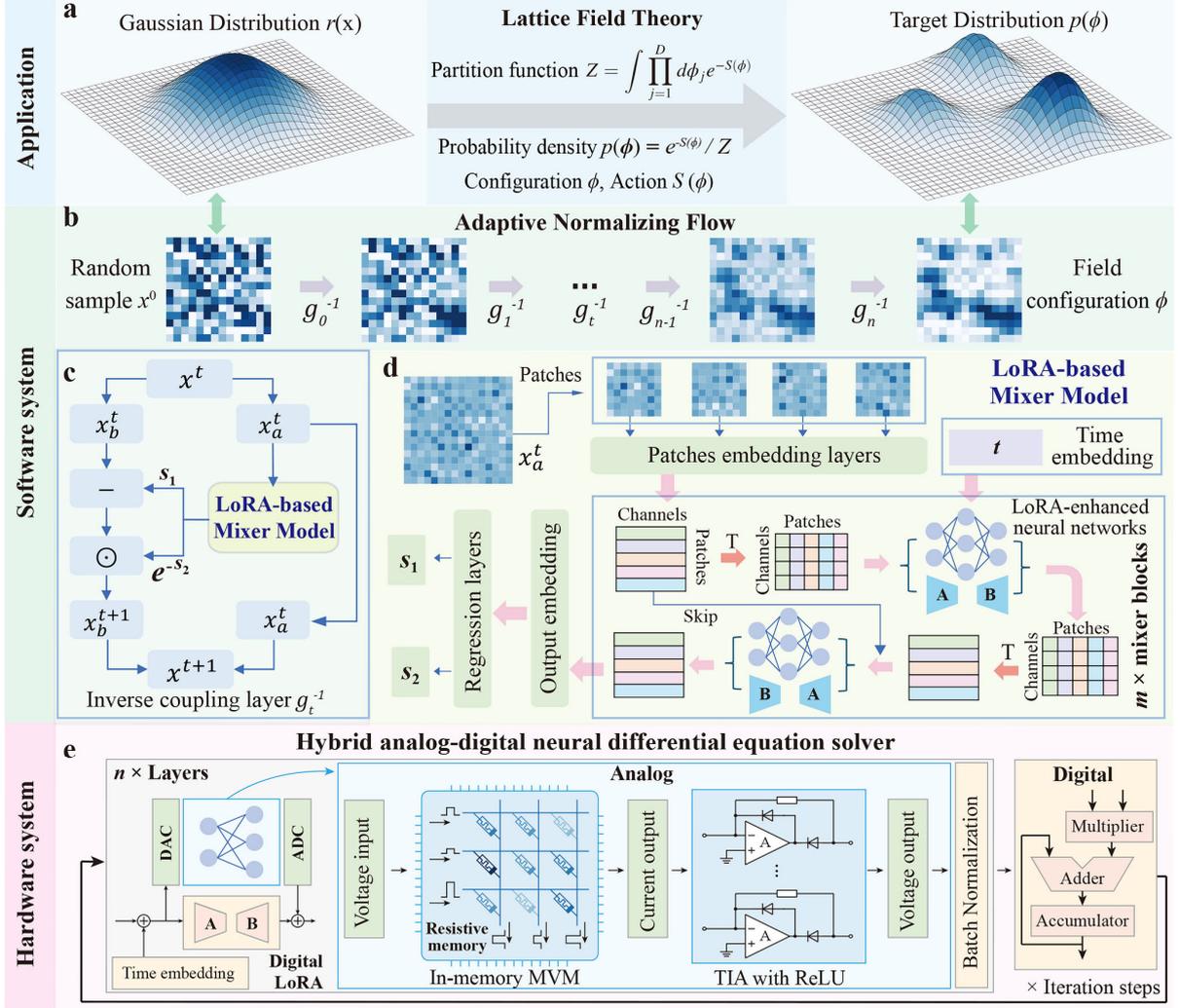

**Fig. 1 | Hardware-software co-design for lattice field configurations generation using resistive memory.** Efficient in-memory computing with resistive memory accelerates the ANF model, enabling high-performance sampling of LFT configurations. **a**, In lattice field theory, the lattice Euclidean action $S(\phi)$ defines the probability density $p(\phi)$ over field configurations $\phi$, providing a physics-informed guide for training ANF models to learn the mapping between Gaussian distributions $r(x)$ and the target distribution $p(\phi)$. **b**, ANF model transforms samples $x$ from a prior Gaussian distribution $r(x)$ to field configurations $\phi$, utilizing invertible coupling layers with shared weights across timesteps. **c**, Structure of the inverse coupling layer $g_t^{-1}$. The input $x^t$ is divided into two subsets $x_a^t$ and $x_b^t$, then $x_a^t$ is processed by the low-rank adaptation (LoRA)-based mixer model to generate $s_1$ and $s_2$, which update $x_b^t$ via subtraction (-) and elementwise multiplication ($\odot$) respectively, followed by recombination

into $x^{t+1}$. **d**, Architecture of LoRA-based mixer model. Inputs are partitioned into patches and embedded into the channel feature space. Within each mixer block, a combination of transpose operations and neural networks enhanced with LoRA is employed to sequentially update features along the patch and channel dimensions. After passing through *m* mixer blocks, output embedding and regression layers produce the final output. Time embeddings are integrated into every mixer block. **e**, Hybrid analog-digital neural differential equation solver based on resistive memory, designed to accelerate the ANF model. Blue blocks denote operations performed using analog computation, while yellow blocks indicate operations executed digitally.

# Results

**Resistive in-memory computing enabled by a hybrid analog-digital neural differential equation solver**

**Fig. 2a-c** depict the hierarchical system structure of the hybrid analog-digital neural differential equation solver, including the resistive memory 1-transistor-1-resistor (1T1R) cell (**a**), the array (**b**), and the system board (**c**). **Fig. 2a** shows the cross-sectional transmission electron microscopy (TEM) image of a 1T1R cell, where a transistor is connected in series with a TiN/Ta$_2$O$_5$/TaO$_x$/TiN resistive memory device, fabricated using 180 nm technology node. The conductance of the resistive memory is governed by oxygen vacancy-based conducting channels in TaO$_x$ layers. **Fig. 2b** displays the schematic and optical image of a resistive memory array composed of 1T1R cells. The resistive memory array serves as the analog matrix multiplier, which, together with a digital Xilinx ZC706 system-on-chip (SoC), constitutes the hybrid analog-digital neural differential equation solver (**Fig. 2c**). **Fig. 2d** shows 200-cycle quasi-static I-V sweeps, demonstrating highly uniform bipolar resistive switching of the resistive memory cell. **Fig. 2e-h** illustrate the analog programming performance of the resistive memory, spanning from single devices to arrays. **Fig. 2e** shows the single-shot set programming of the resistive memory cell, demonstrating that mean target conductance roughly linearly increases with the WL set voltage. **Fig. 2f** displays the iterative programming

of four resistive memory devices, with the conductance converging to the target value after around 20-30 cycles. **Fig. 2g** shows the distribution of programmed conductance values with target values ranging from 20 to 100 µS (in steps of 5 µS), where the std of the distribution peaks is around 0.54. **Fig. 2h** presents the programmed "AI4S" analog conductance pattern on a resistive memory array, as well as the distribution of programming errors, indicating that over 95% of the devices were successfully programmed within a tolerance range of -3.5 µS to 3.5 µS. **Fig. 2i-k** illustrate the in-memory matrix multiplication related performance of resistive memory. **Fig. 2i** displays data retention by repeatedly reading from 20 resistive memory cells with different conductance states (spanning from 30 µS to 95 µS) over a period of $10^6$ s. **Fig. 2j** illustrates the output currents of resistive memory cells (with conductance ranging from 30 µS to 95 µS) under continuous voltage sweeps. Notably, within the input voltage range of -0.1 to 0.1 V, the resistive memory shows highly linear voltage-current relations following Ohm's law. **Fig. 2k** presents an example of matrix multiplication implemented using the resistive memory array. For a fully-connected neural network, the weights are mapped to conductance values ranging from 20 to 80 µS. Each input vector is converted to voltages between -0.1 and 0.1 V. The output current for each column is obtained by summing the currents from the four devices in that column, in accordance with Kirchhoff's current law, yielding one element of the output map. Finally, all output currents are inversely mapped to produce the neural network output.

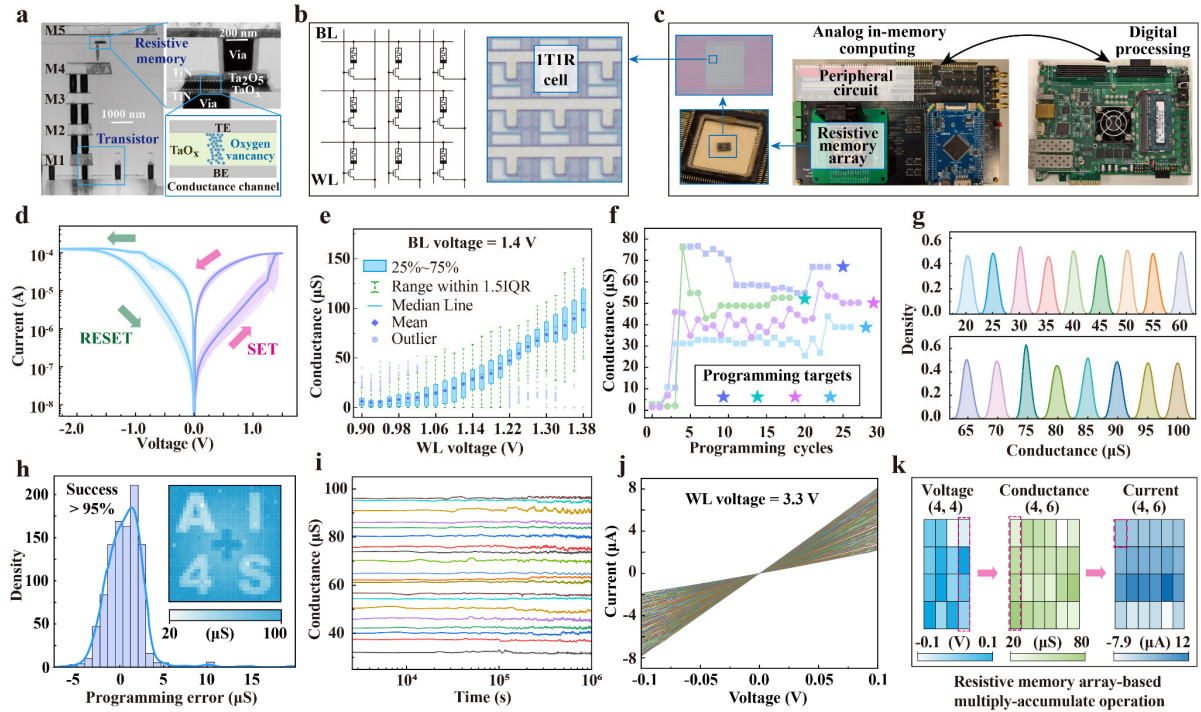

**Fig. 2 | Resistive memory characterization and in-memory matrix multiplication performance in the hybrid analog-digital neural differential equation solver. a**, Cross-sectional transmission electron microscopy (TEM) images of a 1-transistor-1-resistor (1T1R) cell, and a TiN/Ta$_2$O$_5$/TaO$_x$/TiN resistive memory structure with metal vias, alongside a schematic illustrating the formation of conductive channels via oxygen vacancy migration in the TaO$_x$ layer. **b**, Schematic and optical image of an array composed of 1T1R memory cells. **c**, Photograph of a hybrid analog-digital neural differential equation solver based on resistive memory; the analog part utilizes the resistive memory array for efficient in-memory computing, while the digital part consists of a Xilinx ZC706 system-on-chip (SoC). **d**, Quasi-static I-V sweeps of a resistive memory cell, where the solid lines correspond to the average current. **e**, Single-shot set programming of the resistive memory cell to different conductance states using various voltages. **f**, Cyclic programming of several resistive memory cells, with conductance tuned to a target value. **g**, Distribution of programmed conductance values, with target values ranging from 20 to 100 μS (in 5 μS steps). **h**, Programmed "AI4S" analog conductance pattern on a 32 × 32 resistive memory array, and the distribution of programming errors. **i**, Repeated readout of 20 resistive memory cells with different conductance states. **j**, Output currents of resistive memory cells (from 30 to 95 μS) with various conductance states under continuous voltage sweeps ranging from -0.1 to 0.1 V. **k**, Illustration of matrix multiplication using the resistive memory array. Input signals are mapped to voltages (from -0.1 to 0.1 V), and weights are mapped to conductance values (from 20 to 80 μS).

**Experiments on scalar $\phi^4$ lattice field**

We validated our software-hardware co-design on a scalar $\phi^4$ lattice field by generating two-dimensional configurations across various lattice sizes. **Fig. 3a** depicts the workflow for LFT simulation and computation of the corresponding physical observables. The lattice action $S(\phi)$ of the $\phi^4$ theory (**top panel**) defines the target probability density $p(\phi)$ for LFT configurations and informs the physics-guided training of the ANF model. The fully-trained ANF model (**middle panel**) is subsequently implemented on a resistive memory-based hybrid analog-digital neural differential equation solver to generate large ensembles of field configurations $\phi$ and their corresponding probability densities $q(\phi)$ (see **Supplementary Figure 4**). The generated configurations $\phi$ serve as proposals in the MCMC sampling procedure (**bottom panel**). Specifically, the acceptance probability for each configuration is calculated based on its generated probability density $q(\phi)$ and the target probability density $p(\phi)$, determining whether the configuration $\phi$ is accepted or rejected (see **Supplementary Figure 5**). Finally, all the accepted configurations $\phi$ are used to compute physical observables of the $\phi^4$ field (see **Methods**).

**Fig. 3b** displays the autocorrelation function of the average magnetization $M$ for lattice size 4 and 24, demonstrating that the co-design substantially suppresses autocorrelations along the Markov chain compared to the HMC method. The correspond integrated autocorrelation time increases from approximately 9.9 and 23.34 for HMC as the lattice size grows, while for co-design it remains low at approximately 2.1 and 2.8. Despite the variation in autocorrelation time for co-design, which is mainly due to fluctuations in the MCMC acceptance rate, it still yields effectively decorrelated configurations at substantially lower computational cost (see **Supplementary Figure 6**). **Fig. 3c-e** present the physical observables of the $\phi^4$ field computed for various lattice sizes and coupling parameters, using both the HMC method and our co-design. Whereas HMC decorrelates configurations by retaining only one in twenty samples, our co-design directly generates statistically independent configurations for different coupling parameters via LoRA fine-tuning. **Fig. 3c,d** compare the Ising energy and correlation length by both approaches, demonstrating that our co-design can accurately capture the energy density and correlation properties of the system. **Fig. 3e** shows the statistical error in susceptibility as a function of the number of configurations (see **Supplementary Figure 7**). In both the HMC method and our co-design, generating more configurations is required to reduce the statistical error. **Fig. 3f-i** illustrate the performance advantage of our resistive memory-based hybrid analog-digital neural differential equation solver. **Fig. 3f** compares the computational accuracy of ANF implemented in hardware (hybrid analog-digital solver) and software, as well as HMC method in software. While the effective sample size (ESS)[61] is marginally lower for hardware than for software, the susceptibility calculated by hardware ANF closely matches HMC results. **Fig. 3g** shows the distribution of ANF model weights and average MAC operations across analog and digital parts of the solver. For lattice sizes of 4, 6, 12, and 24, the analog resistive memory stores approximately 92.9%, 93.0%, 96.0% and 95.7% weights of ANF model, and performs approximately 94.0%, 95.9%, 98.9% and 99.2% of the total MAC operation counts per inference, respectively, within the hybrid analog-digital solver. These results indicate that more MAC operations of ANF leverage efficient analog in-memory computing as the model scales, with minimal overhead from LoRA-based fine-tuning. **Fig. 3h,i** compare the average inference speed and energy efficiency of our hybrid analog-digital solver and digital-only solver. For single inferences generating configurations with various lattice sizes, the hybrid analog-digital solver achieves speedups of approximately 5.9-, 8.8-, 14.7-, and 16.1-fold (**Fig.**

**3h**) and reduces average energy consumption by 65.9%, 85.3%, 97.4%, and 98.6% (top panel in **Fig. 3i**), respectively. Correspondingly, the average energy efficiency (MAC operations per joule) increases by approximately 2.9-, 6.8-, 38.5-, and 73.7-fold (bottom panel in **Fig. 3i**).

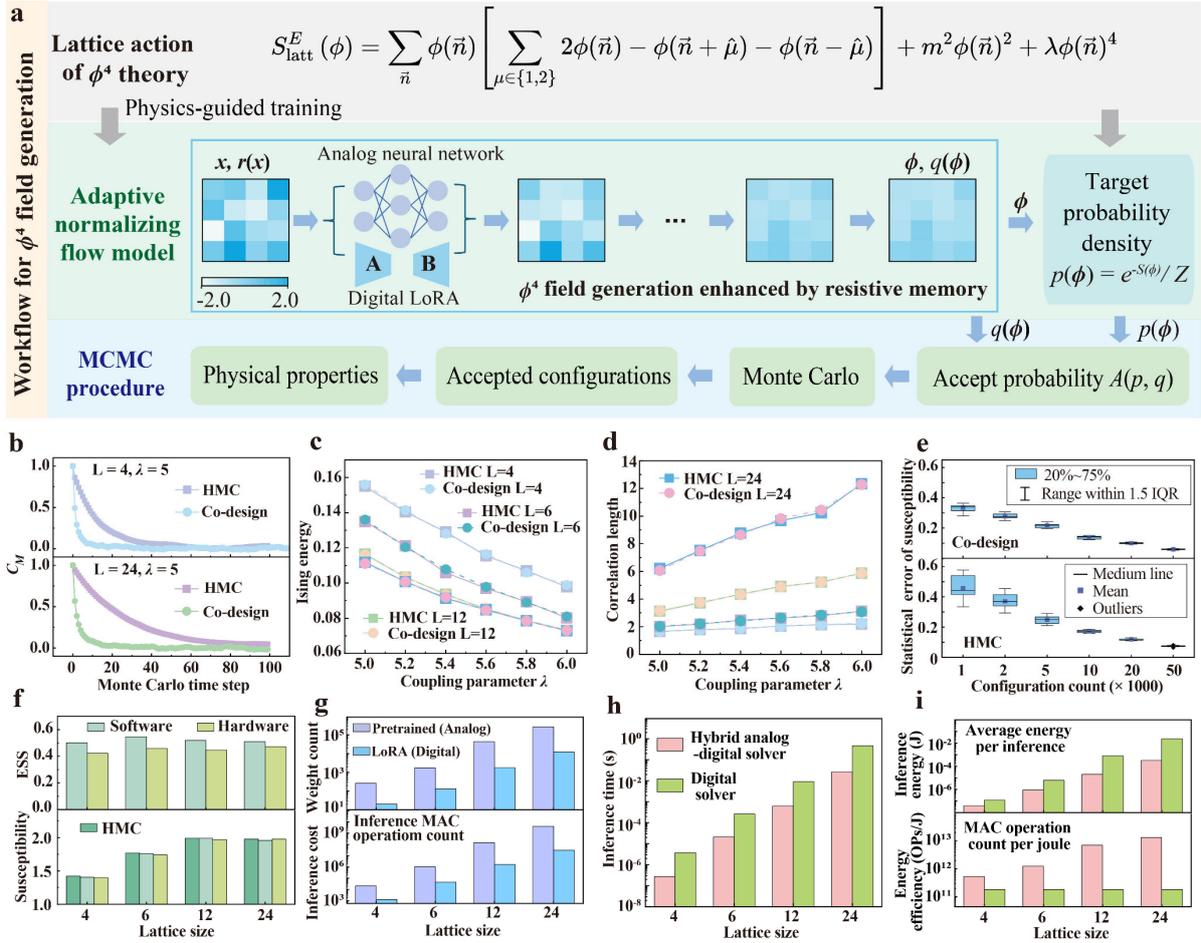

**Fig. 3 | Experimental results of simulating scalar $\phi^4$ theory using the co-design. a,** Schematic workflow of LFT simulation employing ANF model on a resistive memory-based hybrid analog-digital solver for $\phi^4$ theory. The lattice action defines the target probability density $p(\phi)$ and guides ANF model training. Then the ANF is implemented on a hybrid analog-digital solver to generate field configurations $\phi$ with probability densities $q(\phi)$, which are used as proposals in MCMC procedure. **b,** Autocorrelation function of the average magnetization $M$. **c,d,** Comparison of Ising energy (**c**) and correlation length (**d**) computed via HMC and co-design for various lattice sizes and couplings. Experimental results for lattice sizes 4 and 6, as well as simulated results for larger lattices, show excellent agreement with HMC. **e,** Statistical error in susceptibility as a function of configuration count. In box plots, box limits denote the 25th and 75th percentiles, and whiskers extend to the most extreme data

points within 1.5 times the interquartile range (IQR) from the lower and upper quartiles. Data points beyond this range are defined as outliers and are shown as individual dots. The sample size n for each box plot is n = 10 independent measurements per group. **f**, Accuracy comparison of ANF implemented in hardware (hybrid analog-digital solver) and software, as well as HMC method in software. ESS serves as a crucial metric, quantifying the number of effectively independent samples generated by ANF models. **g**, Number of analog-implemented pretrained weight and digital-implemented LoRA weights in the ANF model, and the corresponding average MAC operation count per inference. **h,** Inference speed comparison. The average inference time required to generate a single configuration using the ANF model on the hybrid analog-digital solver and the digital-only solver. **i**, Comparison of average inference energy consumption and energy efficiency (MAC operations per joule).

**Experiments on effective field theory of graphene wire**

We then applied our software-hardware co-design to generated LFT configurations for the EFT of graphene wire[58]. **Fig. 4a** illustrates the workflow, where the graphene wire is represented as a bosonic lattice field and field configurations are generated via our co-design. In low-dimensional graphene wires, electrons and holes are confined to propagate predominantly along the longitudinal direction, with electromagnetic interactions mediated by internal pseudo-photon fields[60]. Within an EFT framework, the dynamics of fermionic (electron and hole) fields and the internal pseudo-photon field are described in 1+1 dimensions, while coupling to external electromagnetic fields is retained. By appropriately rescaling the photon field and electromagnetic coupling, the resulting theory can be mapped onto the Schwinger model[59,62]. In this paradigm, the process of bosonization[63-65] allows the collective excitations in the graphene wire to be represented by a bosonic field, providing a unified description of its low-energy properties. We subsequently train the ANF model using the boson lattice action for the graphene wire, leveraging a resistive memory-based hybrid analog-digital neural differential equation solver to accelerate the generation of field configurations (see **Supplementary Figure 8** and **9**). Compared to HMC, our co-design reduces the integrated autocorrelation time by approximately 13.9 times (see **Supplementary Figure 6**), thereby improving the efficiency of generating independent configurations and reducing computational costs.

**Fig. 4b-e** present the physical observables of the graphene wire field computed for various lattice sizes and action parameters, using both the HMC method and our co-design, demonstrating that our co-design accurately simulates the EFT of graphene wires. **Fig. 4b** shows the distribution of the average field value per configuration, indicating that the configurations generated by our co-design are statistically consistent with those from HMC. **Fig. 4c** shows the zero-momentum two-point correlation function, whose accurate reproduction by our co-design demonstrates reliable capture of the system's temporal correlations and underlying dynamics. **Fig. 4d** presents the mass gap, obtained by fitting the exponential decay of the two-point correlation function, confirming that our co-design can correctly reproduce this key feature about the dispersion relation in the graphene wire. **Fig. 4e** shows the chiral condensate, a measure of the ground-state structure.

**Fig. 4f-i** illustrate the hardware performance of our resistive memory-based hybrid analog-digital neural differential equation solver. **Fig. 4f** compares the computational accuracy of ANF implemented in hardware (hybrid analog-digital solver) and software, as well as HMC method in software. While the ESS is marginally lower for hardware than for software, the absolute magnetization $|M|$ calculated by hardware ANF closely matches HMC results. **Fig. 4g** displays the number of analog- and digital-implemented weights in the ANF model, as well as the average MAC operation count per inference. Like $\phi^4$ theory, for lattice sizes of 6, 12, 24, and 48, the analog resistive memory stores approximately 93.9%, 92.1%, 94.4% and 92.7% weights of ANF model, and performs approximately 98.7%, 98.2%, 98.9% and 99.6% of the total MAC operation counts per inference, respectively, within the hybrid analog-digital solver. **Fig. 4h,i** compare the average inference speed and energy efficiency of our hybrid analog-digital and digital-only solvers. For single inference generating configuration of various lattice sizes, the hybrid analog-digital solver achieves approximately 13.6-, 12.2-, 14.5-, and 17.3-fold faster inference (**Fig. 4h**) and reduces average energy consumption by approximately 93.3%, 95.7%, 97.9%, and 99.3% (top panel in **Fig. 4i**), respectively. Correspondingly, the average energy efficiency (MAC operations per joule) increases by approximately 15.0-, 23.0-, 47.3-, and 138.0-fold (bottom panel in **Fig. 4i**).

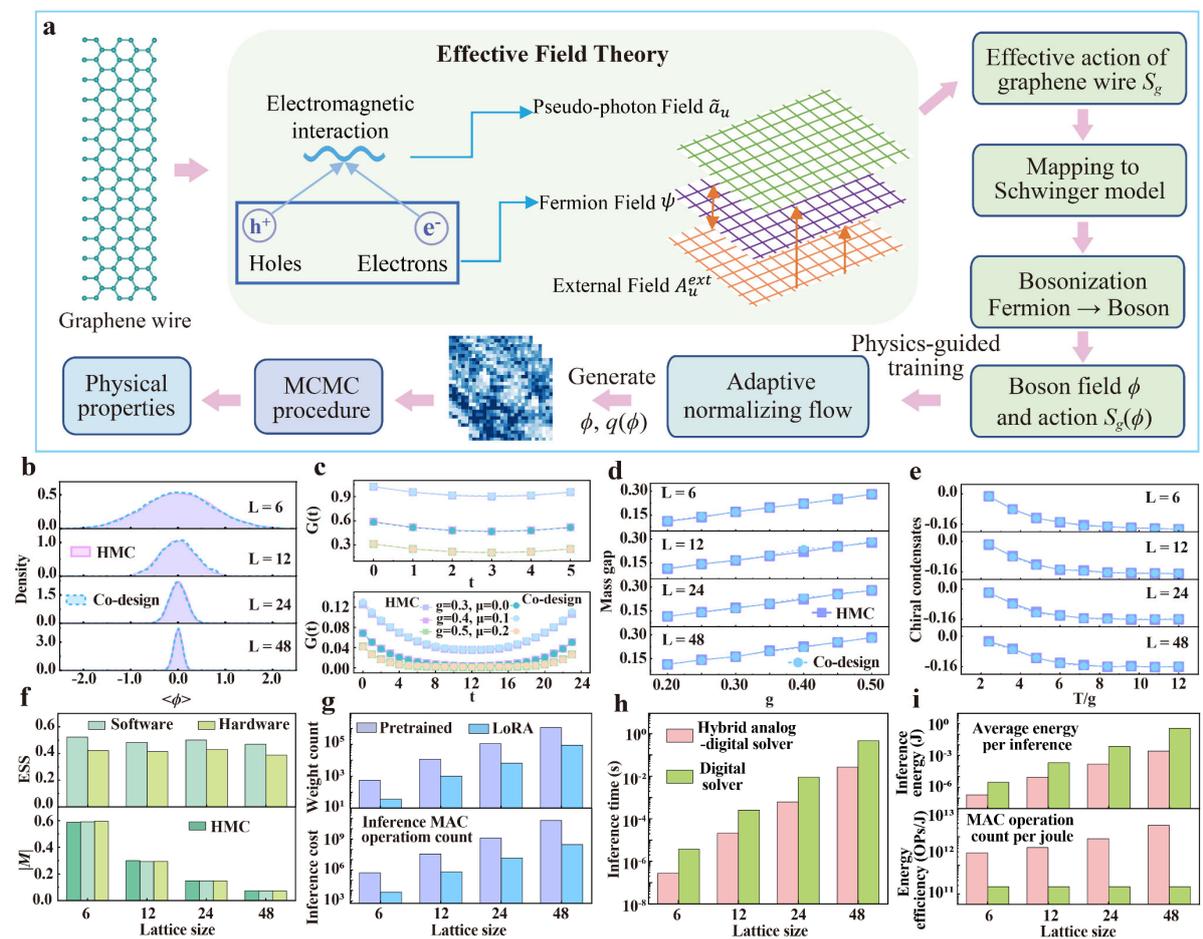

**Fig. 4 | Experimental results of simulating graphene wire effective field theories using the co-design. a,** Schematic illustration of generating graphene wire field configurations and calculating the physical properties. The graphene wire is modeled as a boson lattice field with action $S_g$, which is then mapped onto the Schwinger model. Bosonization yields the bosonic field $\phi$ and its corresponding action $S_g(\phi)$. The action $S_g(\phi)$ is used to train a ANF model, which is further accelerated by a hybrid analog-digital neural differential equation solver based on

resistive memory. The generated configurations serve as proposals in the MCMC procedure, and finally the accepted configurations are utilized for the calculation of physical observables of graphene wire. **b**, Comparison of configuration distributions between co-design (results for lattice size of 6 are measured experimentally; results for larger lattices are obtained via simulation) and HMC across various lattice sizes. **c**, Comparison of zero momentum two-point correlation function. **d**, Mass gaps across various lattice sizes and coupling parameters. **e**, Chiral condensate measurements across various lattice sizes and temperature $T/g = (L_\tau g)^{-1}$. **f**, Accuracy comparison of ANF implemented in hardware (hybrid analog-digital solver) and software, as well as HMC method in software. **g**, Number of analog-implemented pretrained weight and digital-implemented LoRA weights in the ANF model, and the corresponding average MAC operation count per inference. **h,** Inference speed comparison. The average inference time required to generate a single configuration using the ANF model on the hybrid analog-digital solver and the digital-only solver. **i**, Energy efficiency comparison. The ANF model achieves lower average inference energy per sample on the hybrid analog-digital solver compared to the digital-only solver, allowing more MAC operations per joule.

## Discussion

We have developed a resistive memory-based hardware-software co-design to enhance LFT simulation. On the software side, ANF is capable of learning complex target distributions and generating independent configurations in parallel for the MCMC sampling procedure, thus achieving higher computational efficiency while maintaining accuracy for physical observables comparable to that of HMC. Consequently, our co-design demonstrates a critical advantage for large-scale, highly complex physical systems, such as lattice scalar field theories, critical phenomena in statistical mechanics, and models of spontaneous symmetry breaking. Compared to conventional normalizing flow models, our ANF leverages LoRA fine-tuning, enabling rapid and low-cost adaptation to varying LFT parameters. This flexibility enables efficient exploration of a wide parameter space without retraining the entire model, making it highly scalable and adaptable for diverse physical scenarios.

On the hardware side, we have developed a resistive memory-based hybrid analog-digital neural differential equation solver that delivers both faster computation and higher energy efficiency compared to state-of-the-art GPUs. In addition, we integrate digital LoRA to compensate for programming error inherent to resistive memory, achieving computation accuracy comparable to conventional approaches. On this basis, our hybrid analog-digital solver offers superior flexibility and scalability, enabling it to address a wider range of computational tasks. As the system size increases, the performance gains of our hybrid solver become even more pronounced, underscoring its potential for scalable LFT simulations.

Looking ahead, our hardware platform could benefit from advanced fabrication processes and the development of larger-scale resistive memory chips. Such improvements would allow our co-design to address even more extensive LFT simulations, further reducing hardware costs and enhancing scalability. In addition, our co-design could be extended to incorporate additional physical constraints in the ANF model, such as strict gauge symmetries, thereby enabling the generative modeling of complex gauge fields and facilitating simulations in nuclear physics, condensed matter, and other areas of many-body physics.

## Methods

### Lattice field theory

In lattice field theory, the continuous space-time of a quantum field theory is discretized onto a finite, regular lattice, enabling non-perturbative studies of strongly interacting systems via numerical simulation. In this framework, the continuous fields are replaced by discrete variables defined on the sites or links of a D-dimensional lattice. The discretization introduces a lattice spacing $a$, which acts as an ultraviolet regulator, and physical results are obtained in the continuum limit as $a \rightarrow 0$. The dynamics of the field are governed by an action $S(\phi)$, which encodes the interactions and symmetries specific to the underlying physical theory. Field configurations $\phi$ are sampled according to the Boltzmann weight $e^{-S(\phi)}$, leading to the probability distribution:

$$p(\phi) = \frac{1}{Z} e^{-S(\phi)} \tag{1}$$

$$Z = \int \prod_{j=1}^{D} d\phi_j e^{-S(\phi)} \tag{2}$$

where $j$ indexes the $D$ components of $\phi$, $S(\phi)$ is the action that defines the theory, and $Z$ is the partition function ensuring normalization.

**Adaptive normalizing flow**

The ANF model incorporates multiple inverse coupling layers $g_t^{-1}$ that share weights across different timesteps, with $t$ representing distinct time embeddings. The process begins by sampling a random variable $x$ from a Gaussian distribution $r(x)$. Each inverse coupling layer then applies a reversible transformation to $x$:

$$x^{t+1} = g_t^{-1}(x^t) \tag{3}$$

Within each inverse coupling layer, a mask partitions the input matrix $x^t$ into two subsets $x_a^t$ and $x_b^t$:

$$M_{i,j}^t = (t \bmod 2) \text{ if } ((i+j) \bmod 2 = 0) \text{ else } (1-(t \bmod 2)) \tag{4}$$

$$x_a^t = x^t \circ M^t \tag{5}$$

$$x_b^t = x^t \circ (1 - M^t) \tag{6}$$

where $\circ$ denotes the elementwise multiplication. The transformation is applied only to $x_b^t$, while the $x_a^t$ remains unchanged (the "frozen" part). The transformation of the $x_b^t$ typically depends on the values of the frozen $x_a^t$.

As illustrated in **Fig. 1c**, the frozen subset $x_a^t$ is processed by the LoRA-based mixer model

to predict $s_1$ and $s_2$. Subsequently, a new $x^{t+1}$ is computed based on these predictions:

$$x^{t+1} = (x_b^t - s_1) \circ e^{-s_2} + x_a^t \tag{7}$$

Meanwhile, the Jacobian matrix for such an inverse coupling layer is given by:

$$J = \left| \det_{kl} \frac{\partial [g_t(x)]_k}{\partial x_l} \right| = \prod_k e^{(s_2)_k}, \tag{8}$$

where $k$ runs over the components in $s_2$.

Across various timesteps, the roles of $x_a^t$ and $x_b^t$ are alternated to ensure both subsets are transformed. The invariance of the frozen subset and the dependence of the transformation on this subset together guarantee the invertibility of the overall operation. The final output of the ANF model includes configuration $\phi$ and probability density $q(\phi)$:

$$q(\phi) = r(x) \prod_t J_t^{-1} \tag{9}$$

The training objective of the ANF model is to ensure that the generated distribution $q(\phi)$ closely approximates the target distribution $p(\phi)$. The training process can be initiated solely with the action $S(\phi)$, without the need for additional datasets. The loss function used is a shifted Kullback-Leibler (KL) divergence between the target distribution $p(\phi)$ and the proposal distribution $q(\phi)$ generated by the ANF model:

$$\mathcal{L} = D_{KL}[p(\phi) \| q(\phi)] = \frac{1}{N} \sum_{i=1}^{N} [\log p(\phi_i) - \log q(\phi_i)] \tag{10}$$

During training, a key metric used to evaluate ANF's performance is the effective sample size (ESS):

$$ESS = \frac{1}{N} \frac{(\sum_{i=1}^{N} p(\phi_i)/q(\phi_i))^2}{\sum_{i=1}^{N} (p(\phi_i)/q(\phi_i))^2} \tag{11}$$

Upon completion of training, the configurations and probability densities generated by the ANF are employed to facilitate the MCMC sampling process. The MH algorithm consists of proposing an updated configuration $\phi'$ to the current configuration $\phi^{i-1}$ and accepting or rejecting the configuration with probability

$$A(\phi^{i-1}, \phi') = \min\left(1, \frac{q(\phi^{i-1})}{p(\phi^{i-1})} \frac{p(\phi')}{q(\phi')}\right) \qquad (12)$$

**Architecture of LoRA-based mixer model**

The LoRA-based mixer model is composed of trainable patch embedding layers, mixer blocks, output embedding layers, and regression layers. In this architecture, the LoRA-based neural network is primarily applied to the mixer blocks[66]. For a given $L_x \times L_y$ lattice field, the input feature is first divided into several patches, each of size $P \times P$, resulting in a total of $S = \frac{L_x \times L_y}{P \times P}$ patches.

In the patch embedding layers, each patch is transformed into a feature representation within a hidden channel dimension $C$ through a trainable neural network (either fully-connected layers or selectable convolutional layers), forming the input matrix $X \in R^{S \times C}$,

$$X = \text{Embedding}(\text{Patches}) \qquad (13)$$

which is a two-dimensional matrix where rows correspond to spatial locations (tokens) and columns correspond to channel features.

The core of the mixer model comprises the mixer blocks, each of which alternates between two types of multi-layer perceptron (MLP) blocks: a token-mixing MLP and a channel-mixing MLP. Each block processes either the spatial or the channel dimension independently, and integrates information via residual (skip) connections and layer normalization. In the token-mixing MLP, information is exchanged between different spatial positions (patches). Mathematically, for the $i$-th channel, the operation can be expressed as:

$$U_{*,i} = X_{*,i} + W_2 \sigma (W_1 \text{BatchNorm}(X+t)_{*,i}), \text{ for } i = 1...C \tag{14}$$

where $t$ denotes the time embedding, $(X+t)_{*,i}$ denotes the matrix representing the $i$-th channel's elements across all spatial positions (i.e., fixing channel and traversing all spatial locations), $W_1 \in R^{D_S \times S}$ and $W_2 \in R^{S \times D_S}$ represent the weights of fully-connected layers, $D_S$ represents the hidden layer width, $\sigma$ represents the ReLU activation function.

To enable fine-tuning across different LFT parameters, LoRA layers can be introduced in parallel with $W_1$ and $W_2$, providing an elegant and efficient pathway for model adaptation without extensive retraining. Each LoRA module consists of two fully connected layers: layer A projects the input into a low-dimensional space, while layer B restores this low-dimensional feature back to the original dimension. The output of these LoRA layers is then added to the corresponding base network's output, enabling flexible and lightweight adaptation of the model parameters.

In the channel-mixing MLP, information is exchanged between different channel features at each spatial position. For the $j$-th spatial position, the operation is given by:

$$Y_{j,*} = U_{j,*} + W_4 \sigma (W_3 \text{BatchNorm}(U)_{j,*}), \text{ for } i = 1...S. \tag{15}$$

where $U_{j,*}$ represents the feature vector of all channels at the $j$-th spatial position. Similarly, both $W_3$ and $W_4$ are enhanced with LoRA layers.

After passing through $m$ mixer blocks, the output $Y^m$ is fed into the trainable output embedding layers, where the matrix $Y^m$ is reshaped as:

$$\text{Reshape } (Y^m \in R^{S \times C}) \rightarrow Y_r^m \in R^{\frac{H}{P} \times \frac{W}{P} \times C} \tag{16}$$

Then, a transposed convolution is applied to restore the feature map to the same spatial dimensions as the original input:

$$Y_e \in R^{H \times W \times 2} = \text{TransposedConv}(Y_r^m) \tag{17}$$

Finally, in the regression layers, the per-pixel fully-connected (Per-pixel FC) operation is implemented, which processes each pixel individually across the spatial dimensions while maintaining the overall spatial structure of the feature map:

$$O \in R^{H \times W \times 2} = \text{Per-pixel FC}(Y_e) \tag{18}$$

The per-pixel FC is primarily realized using a combination of transposition operations and fully connected layers. Finally, the output is finally split into $s_1$ and $s_2$:

$$s_1 = O_{:,:,0} \in R^{H \times W} \tag{19}$$

$$s_2 = O_{:,:,1} \in R^{H \times W} \tag{20}$$

**Fabrication of resistive memory chips**

The resistive memory chip utilized in this study was fabricated using a 180 nm technology node and comprises a 32 × 32 1T1R array. Stacked-layer memory cells were constructed atop the drain terminals of the transistors and integrated between the Metal4 and Metal5 layers during the back-end-of-line process. Each memory cell features TiN as both the bottom (BE) and top electrodes (TE), encapsulating a dual-layer $Ta_2O_5/TaO_x$ dielectric. Specifically, a 700 nm diameter via was defined in the Via4 layer using photolithography and etching, and subsequently filled with a 40 nm TiN layer via physical vapor deposition (PVD). The 60 nm dual $Ta_2O_5/TaO_x$ dielectric stack was deposited by atomic layer deposition (ALD), followed by the formation of a 40 nm TiN top electrode and the deposition of the top Metal5 layer.

The 1T1R cells within each row share a Word Line (WL) and Source Line (SL), connecting to the gate and source terminals of the transistors, respectively. Cells in the same column share a Bit Line (BL), which interfaces with the TE terminals, thus forming a crossbar array. To enhance device performance, the chips underwent post-fabrication annealing at 400 °C for 30 minutes under vacuum, resulting in devices exhibiting high endurance and reliability.

**The hybrid analog-digital neural differential equation solver**

As shown in **Fig. 2c**, the hybrid analog-digital system combines a 180-nm resistive memory computing-in-memory chip and a Xilinx ZC706 SoC on a printed circuit board. This integration enables the system to handle both analog and digital conversions seamlessly. In both the experiments, the speed and energy efficiency of the digital part are estimated on the basis of the performance parameters of the baseline GPU, ensuring a fair comparison. To generate 64-way parallel analog voltages, the system utilizes an eight-channel digital-to-analog converter (DAC80508, Texas Instruments) with a 16-bit resolution. For vector-matrix multiplication, the resistive memory chip receives a d.c. voltage through a four-channel analog MUX (CD4051B, Texas Instruments) connected to an 8-bit shift register (SN74HC595, Texas Instruments). The multiplication result is carried by the current from the source line, which is then converted to voltages using trans-impedance amplifiers (OPA4322-Q1, Texas Instruments). These voltages are subsequently read by an analog-to-digital converter (ADS8324, Texas Instruments, 14-bit resolution) and sent to the Xilinx SoC for further processing.

**Deploying ANF on a hybrid analog-digital neural differential equation solver**

Withing the hybrid analog-digital solver, the resistive memory-based analog part manages the base neural network weights, while the digital part handles a smaller set of LoRA weights and operations such as normalization. Specifically, in the LoRA-based mixer model, the fully connected and convolutional layers, as well as the ReLU activation functions within the patch embedding, output embedding, and output regression layers, are implemented by the analog component, while the normalization layers (such as batch normalization) in these modules are managed by the digital component. Within each mixer block, the large-scale base neural network is realized by the analog component, whereas the compact LoRA modules and time embedding are executed by the digital component. Finally, at each time step of the inverse coupling layer, the sample update from $x^t$ to $x^{t+1}$ is likewise computed by the digital part. In our experiments, configurations with small lattice sizes, specifically $L = 4$ and $L = 6$ in the $\phi^4$ field theory and $L = 6$ in the EFT of the graphene wire, were generated using a 32×32 resistive memory array. For larger lattice sizes, configurations were obtained through simulations based on hardware parameters.

**Hardware performance estimation**

To estimate the energy consumption of the resistive memory-based hybrid analog-digital neural differential equation solver, we counted the number of MAC operations through the ANF models and separately estimated the energy consumption of single MAC operations for both analog and digital parts. The energy consumption of the digital part is 549 fJ per MAC operation, which is the same as a state-of-the-art GPU. The energy consumption of the analog part includes the resistive memory array and periphery circuit, with the performance of the peripheral circuits benchmarked against a state-of-the-art reference system. The energy consumption of the resistive memory array is estimated with an input read voltage of 0.2 V, a read pulse width of 10 ns, and an average conductance of 50.00 μS, resulting in 2.0 fJ per cell per multiplication. The energy consumption of the peripheral circuitry primarily includes TIA, digital-to-analog conversion (DAC) and analog-to-digital conversion (ADC). In our hybrid analog-digital solver, the energy consumption per operation of the DAC is estimated to be 0.227 pJ. During a single vector-matrix multiplication, the average energy consumption for the DAC is given by the product of the per-operation energy and the length of the input vector. The TIA exhibits an energy consumption of 3 pJ per operation; correspondingly, its average energy consumption during a vector-matrix multiplication is determined by multiplying the per-operation energy by the length of the output vector. For the ADC, the energy consumption per operation is estimated to be 11.3 pJ, and its average energy consumption for a vector-matrix multiplication is similarly the product of the per-operation energy and the length of the output vector.

To estimate the average computational speed of the resistive memory-based hybrid analog-digital neural differential equation solver and its digital counterpart, we scale both systems to a unit area (ensuring identical chip areas of analog and digital parts) and evaluate the area efficiency, defined as the time per MAC operation. For the analog part, we construct a 32 × 32 crossbar array based on a 1T1R structure. The estimated total time required for the resistive memory chip to complete a matrix calculation includes the resistive memory read pulse (10 ns), MUX decoder (6.1 ns), ADC (0.8 ns, two instances), and shifter & adder (0.19 ns). With an array area of 0.01 mm², the resulting area efficiency of the analog core is approximately 5.72 TFLOPs/mm². For the digital part, we directly adopt the computational results from an Nvidia RTX5090 GPU, which achieves an area efficiency of 0.14 TFLOPs/mm². During inference on the hybrid analog-digital solver, for each network layer, we calculate the computational time of both the analog and digital parts based on the MAC operation count, and take the slower of the two as the final computation time for that layer. The sum of the computation times across all layers gives the average inference time. For the digital-only solver running the ANF model, the computation time for each layer is directly calculated using the area efficiency of 0.14 TFLOPs/mm².

**Scalar $\phi^4$ field theory**

The action for a real scalar $\phi^4$ field theory in two dimensions is given by

$$S(\phi) = \int d^2x (\partial_\mu \phi)^2 + m^2\phi^2 + \lambda\phi^4 \tag{21}$$

where $\phi$ is a real scalar field, $m^2$ denotes the bare mass (set to -4), and $\lambda$ is the quartic coupling constant (ranging from 5 to 6). To study this theory numerically, the action is discretized on a two-dimensional square lattice $L_x \times L_y$ with lattice space $a$ set to 1,

$$S^E_{latt}(\phi) = \sum_{\vec{n}} \phi(\vec{n}) \left[ \sum_{\mu \in \{1,2\}} 2\phi(\vec{n}) - \phi(\vec{n}+\hat{\mu}) - \phi(\vec{n}-\hat{\mu}) \right] + m^2\phi(\vec{n})^2 + \lambda\phi(\vec{n})^4 \tag{22}$$

where $\phi(\vec{n})$ is defined only at the lattice sites $\vec{n} = (n_x, n_y)$, and the total volume is $V = L_x L_y$. Periodic boundary conditions are imposed in both directions, so that $\phi(L_x, y) \equiv \phi(0, y)$, and similarly for the direction $y$.

The connected two-point correlation function is expressed as

$$C(x) = \frac{1}{V} \sum_y (\langle \phi(y)\phi(y+x) \rangle - \langle \phi(y) \rangle \langle \phi(y+x) \rangle) \tag{23}$$

And its momentum-space representation is represented by

$$C(\vec{p}, t) = \frac{1}{L^{d-1}} \sum_{\vec{x}} e^{i\vec{p}\cdot\vec{x}} C(\vec{x}, t), \tag{24}$$

Then the two-point susceptibility, average Ising energy density[30] and correlation length can be estimated via

$$\chi_2 = \sum_x C(x). \tag{25}$$

$$E = \frac{1}{d} \sum_{1 \leq \mu \leq d} C(\hat{\mu}) \tag{26}$$

$$\frac{1}{\xi} = \frac{1}{L-1} \sum_{x_2=1}^{L-1} \text{arcosh}\left( \frac{C(x_2+1) + C(x_2-1)}{2C(x_2)} \right). \tag{27}$$

The average magnetization is given by

$$M = \left\langle \frac{1}{V} \sum_{\vec{n}} \phi(\vec{n}) \right\rangle \tag{28}$$

The magnitude of the magnetization |M|, represents the absolute value of this average. And the

autocorrelation function of an observable $O$ is given by

$$C_O(t) = \left\langle \left(O_{t_0} - \left\langle O_{t_0} \right\rangle\right)\left(O_{t_0+t} - \left\langle O_{t_0+t} \right\rangle\right) \right\rangle = \left\langle O_{t_0} O_{t_0+t} \right\rangle - \left\langle O_{t_0} \right\rangle\left\langle O_{t_0+t} \right\rangle \tag{29}$$

It measures how the observable $O$ at time $t_0$ correlates with its value at a later time $t_0 + t$. Then the integrated autocorrelation time is given by

$$\tau_{X,\text{int}} = \frac{1}{2} + \frac{1}{C_O(0)} \sum_{t=1}^{T} C_O(t) \tag{30}$$

**Effective field theory of graphene wire**

Graphene wire, owing to its pronouncedly anisotropic geometry, serve as a paradigmatic platform for exploring low-dimensional quantum electrodynamics. In such systems, electrons and holes are confined to propagate predominantly along the wire's longitudinal direction, with electromagnetic interactions mediated by pseudo-photon fields within the wire[58]. The pronounced separation of scales-where the transverse dimension is much smaller than the longitudinal extent-naturally allows the energy spectrum to be partitioned into high-energy transverse modes and low-energy longitudinal modes. The framework of effective field theory[67] exploits this separation by focusing on low-energy degrees of freedom, rendering the details of high-energy (ultraviolet) physics largely irrelevant for the description of long-wavelength phenomena. Within this EFT approach, the dynamics of electrons, holes, and the internal pseudo-photon field are formulated in 1+1 dimensions, while retaining coupling to external electromagnetic fields in the ambient 3+1 dimensional space. The resulting effective action consists of a free Dirac term for the fermions (electrons and holes), a kinetic term for the photon field, and a gauge-invariant coupling achieved via the covariant derivative[58,59]:

$$S \equiv \int d^2\tilde{x} \left[ \bar{\psi} i \tilde{D} \psi + m(\Lambda)\bar{\psi}\psi - \frac{c(\Lambda)}{4} f_{\mu\nu} f^{\mu\nu} \right] \tag{31}$$

$$f_{\mu\nu} = \tilde{\partial}_\mu \tilde{a}_\nu - \tilde{\partial}_\nu \tilde{a}_\mu = \partial_\mu a_\nu - \partial_\nu a_\mu \tag{32}$$

$$\tilde{D}_\mu = \tilde{\partial}_\mu + ie\tilde{a}_\mu + ie\tilde{A}_\mu^{\text{ext}} \tag{33}$$

where $\psi$ is the fermion-field operator, $\tilde{a}_\mu$ is the quantized dynamical field which describes the photons inside the wire, $\tilde{A}_\mu^{ext}$ is a classical external field.

Remarkably, this effective action can be formally mapped onto the Schwinger model[62], a well-known exactly solvable theory of quantum electrodynamics in 1+1 dimensions. The mapping is accomplished by rescaling the pseudo-photon field and the electromagnetic coupling such that the interaction structure matches that of the Schwinger model. By setting $\sqrt{c(\Lambda)}\tilde{a}_\mu \to \tilde{a}_\mu$ and $g = \dfrac{e}{c(\Lambda)}$, the covariant derivative is defined as

$$\tilde{D}_\mu = \tilde{\partial}_\mu + ig\tilde{a}_\mu + ie\tilde{A}_\mu^{ext} \tag{34}$$

Therefore, the quantum electrodynamics of electrons and holes inside the wire is described by the effective action:

$$S_{Sch} = \int d^2\tilde{x}\left[\bar{\psi}\tilde{D}i\psi - m\bar{\psi}\psi - \frac{1}{4}f_{\mu\nu}f^{\mu\nu}\right], \quad \mu,\nu = 0,1 \tag{35}$$

By invoking the bosonization technique developed for the Schwinger model[63-65], the effective action describing the graphene wire can be systematically transformed into an equivalent bosonic representation:

$$S_g(\phi) = \int_x \left\{\frac{1}{2}\partial^\mu\phi\,\partial_\mu\phi - \frac{(g\phi)^2}{2\pi} - u\cos(2\sqrt{\pi}\phi)\right\}, \tag{36}$$

where the parameter $u$ linearly related to the fermion mass. To facilitate LFT simulation, we discretize the action on a lattice (a = 1), yielding the corresponding lattice action:

$$S_g(\phi) = \sum_{\tau=0}^{L_\tau-1}\sum_{x=0}^{L_x-1} \frac{1}{2}(\partial_\tau\phi_{x,\tau})^2 + \frac{1}{2}(\partial_x\phi_{x,\tau})^2 - \frac{(g\phi_{x,\tau})^2}{2\pi} - u\cos(2\sqrt{\pi}\phi_{x,\tau}) \tag{37}$$

Then the zero momentum two-point correlation function can be computed using equations (23) and (24). Subsequently, it is typically fitted to an exponential decay form,

$$G_c(0,t) \sim e^{-m_{\text{gap}}t} \tag{38}$$

allowing for the extraction of the mass gap. The chiral condensate is computed by:

$$\langle \psi \overline{\psi} \rangle = -\frac{e^{\gamma}}{2\pi^{3/2}} gO(1/g) \langle \cos(2\sqrt{\pi}) \rangle \tag{39}$$

with $O(1/g) \approx 10/g$.

# Data availability

Source Data are provided with this paper.

# Code availability

The code used in the current study is available at GitHub (https://github.com/hustmeng/ANF).

# Acknowledgements

This research is supported by the National Key R&D Program of China (Grant No. 2022ZD0117600), the National Natural Science Foundation of China (Grant Nos. 62122004, 62374181, 61821091), the Strategic Priority Research Program of the Chinese Academy of Sciences (Grant No. XDB44000000), Beijing Natural Science Foundation (Grant No. Z210006), Hong Kong Research Grant Council (Grant Nos. 27206321, 17205922, 17212923). This research is also partially supported by ACCESS - AI Chip Center for Emerging Smart Systems, sponsored by Innovation and Technology Fund (ITF), Hong Kong SAR.

# Competing interests

The authors declare no competing interests.